\documentclass[3p,twocolumn,times]{elsarticle} 

\usepackage{lineno,url,hyperref}
\modulolinenumbers[5]

\usepackage{textcomp}
\usepackage[utf8]{inputenc} 
\usepackage[T1]{fontenc}    

\usepackage{enumitem}

\usepackage{color}
\usepackage{amsmath}
\usepackage{amssymb}
\usepackage{mathtools}  
\usepackage{amsthm}
\usepackage{amsfonts}
\usepackage{comment}

\usepackage{nicefrac}       
\usepackage{microtype}      
\usepackage{xcolor}         
\usepackage{algorithm}
\usepackage{bm}
\usepackage{algpseudocode} 
\usepackage{amsthm,amsmath,amssymb,amsfonts}
\usepackage{graphicx}

\usepackage{booktabs}
\usepackage{multirow}
\usepackage{float} 
\usepackage{graphicx} 
\usepackage{xcolor}
\graphicspath{{Figures/}}

\usepackage{caption}  
\usepackage{array}  

\usepackage{comment}

\newcounter{ToDo}
\newcounter{gaocomm}
\newcounter{wangcomm}
\newcounter{Note1}
\definecolor{blue-violet}{rgb}{0.00,0.67,0.80}
\definecolor{mygreen}{rgb}{0.0, 0.5, 0.0}
\definecolor{awesome}{rgb}{1.0, 0.13, 0.32}
\definecolor{bostonuniversityred}{rgb}{1.0, 0.0, 0.0}

\begin{document}

\begin{frontmatter}
\title{{\bf Graph Contrastive Learning with Implicit Augmentations}}

\author[1,2]{Huidong Liang\corref{cor1}\fnref{fn1,fn2}}
\ead{hlia0714@uni.sydney.edu.au}
\author[2]{Xingjian Du\fnref{fn1}}
\ead{duxingjian.real@bytedance.com}
\author[2]{Bilei Zhu}
\ead{zhubilei@bytedance.com}
\author[2]{Zejun Ma}
\ead{mazejun@bytedance.com}
\author[3]{Ke Chen}
\ead{knutchen@ucsd.edu}
\author[1]{Junbin Gao}
\ead{junbin.gao@sydney.edu.au}

\cortext[cor1]{Corresponding author}
\fntext[fn1]{Both authors contributed equally.}
\fntext[fn2]{Work finished during an internship at ByteDance AI Lab.}

\address[1]{Discipline of Business Analytics, The University of Sydney Business School, The University of Sydney, Australia}
\address[2]{ByteDance AI Lab, Shanghai, China}
\address[3]{University of California San Diego, San Diego, United States}

\begin{abstract}
Existing graph contrastive learning methods rely on augmentation techniques based on random perturbations (e.g., randomly adding or dropping edges and nodes). Nevertheless, altering certain edges or nodes can unexpectedly change the graph characteristics, and choosing the optimal perturbing ratio for each dataset requires onerous manual tuning. In this paper, we introduce \textit{Implicit Graph Contrastive Learning} (iGCL), which utilizes augmentations in the latent space learned from a Variational Graph Auto-Encoder by reconstructing graph topological structure. Importantly, instead of explicitly sampling augmentations from latent distributions, we further propose an upper bound for the expected contrastive loss to improve the efficiency of our learning algorithm. Thus, graph semantics can be preserved within the augmentations in an intelligent way without arbitrary manual design or prior human knowledge. Experimental results on both graph-level and node-level tasks show that the proposed method achieves state-of-the-art performance compared to other benchmarks, where ablation studies in the end demonstrate the effectiveness of modules in iGCL.
\end{abstract}

\begin{keyword} graph neural networks \sep contrastive learning \sep latent augmentations \sep graph auto-encoders
\MSC[2010] 00-01\sep  99-00
\end{keyword}
\end{frontmatter}

\mbox{}\vspace{1ex}

\section{Introduction}
Graph Neural Networks (GNN) and their variants \cite{kipf2017semi,velivckovic2018graph,hamilton2017inductive,xu2019powerful} have achieved state-of-the-art performance on both graph-level and node-level tasks such as social network analysis \cite{xu2019link}, molecular interactions \cite{gasteiger2021gemnet} and recommender systems \cite{ying2018graph}. In many scenarios, training end-to-end GNN models with supervision is impractical as label information is frequently unavailable or difficult to retrieve. Meanwhile, Self-Supervised Learning (SSL), an unsupervised method that first trains model on auxiliary tasks without label information and then uses the learned embeddings for downstream tasks, has gradually gained popularity in recent literature \cite{liu2022graph}. As one of the effective pre-training designs, Graph Contrastive Learning (GCL), which aims to maximize the mutual information of the inputs and their counterparts, becomes a current representative of graph SSL method \cite{xie2022self}. 

To generate different views (augmentations) for the input graphs, existing GCL methods at both node-level \cite{velivckovic2019deep,zhu2020deep,zhu2021graph} and graph-level \cite{you2020graph, sun2020infograph, xu2021infogcl} rely on augmentations based on random perturbations, including randomly adding or dropping the graph's edges or nodes, and shuffling or masking node attributes. These augmentation approaches are based on a strong assumption that a small random perturbation will not alter the semantic property of the original graph \cite{you2020graph}. Nevertheless, this assumption involves two critical issues. First, manipulations of edges and nodes on graphs in specific domains can ruin the data. For example, removing or appending certain 
edges or nodes in a molecular graph or a knowledge graph may immediately destruct the fundamental properties of graph \cite{lee2022augmentation}, hence leading to fallacious views for contrastive learning. Second, the level of perturbation is regarded as a very sensitive hyper-parameter in such methods. Ablation studies in existing GCL literature \cite{zhu2020deep, you2020graph, zhu2021graph} indicate that the performance of GCL models based on random augmentations are sensitive to the level of perturbation, where the optimal ratios often need to be tailored for different datasets after extensive hyper-parameter selection process. Therefore, to improve the effectiveness and efficiency in GCL, it is important to find an alternative way for graph data augmentation that can automatically generate different views and also retain graph semantics.

\begin{figure}[t]
\centering
\includegraphics[width=0.98\columnwidth]{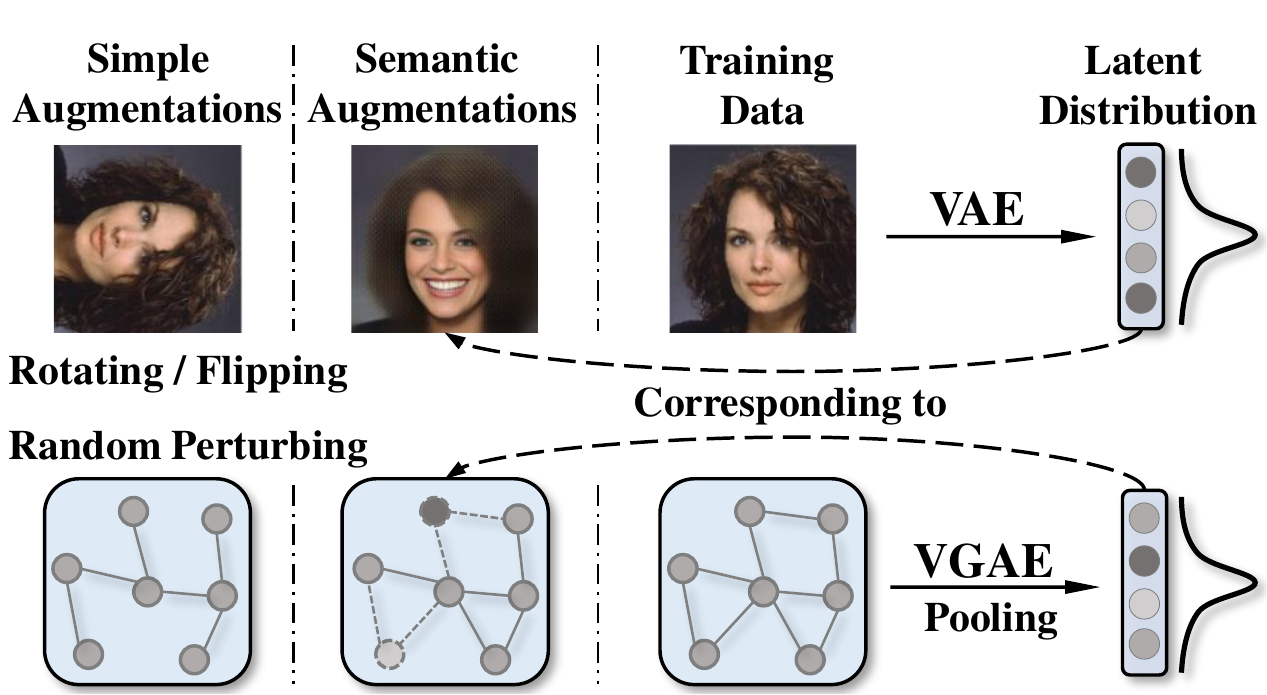} 
\caption{Illustration of simple augmentation and semantic augmentation in terms of image (top) and graph (bottom), where input data is sent to either VAE or VGAE to generate a latent distribution of low-dimensional vector that contains high-level meanings. Then the semantic augmentations of image or graph are performed by sampling from such distributions.} 
\label{fig intro}
\end{figure}

Meanwhile, implicit data augmentation methods \cite{wang2019implicit,li2021metasaug} have shown impressive results over the traditional augmentation methods in computer vision. Inspired by the high-level semantics preserved in deep neural networks' embeddings \cite{lecun2015deep}, these approaches perform augmentation in the latent space rather than using simply flipped or rotated images as augmentations in the input space (illustrated in Figure~\ref{fig intro}), which provides a direction to improve the quality of augmentation in GCL. However, existing implicit augmentation methods are dependent on label information (i.e. supervised) and remain unexplored under graph settings. On the other hand, Variational Auto-Encoder (VAE) \cite{kingma2014auto} is another powerful SSL model trained by reconstructing input features. Compared to simple image augmentation methods, embeddings sampled from the latent distributions in VAE correspond to high-level semantic meanings in the original image \cite{Liu2020towards}. As demonstrated in Figure \ref{fig intro}, semantic augmentations can be performed in the latent space of VAE or VGAE for different formats of data.

Despite of the reconstruction power of VAE and VGAE, the latent distributions often overlap with each other, making the embeddings difficult to separate for downstream tasks \cite{mathieu2019disentangling}. Therefore, we seek to address this overlapping issue by using graph contrastive method that aims to separate embeddings in the representation space, and in return, the semantic latent augmentations from VGAE can further improve the effectiveness of GCL, leading to our innovations below.

\subsubsection*{Contributions} In this paper, we introduce \textit{Implicit Graph Contrastive Learning} (iGCL) for both node-level and graph-level tasks, which exploits semantic latent augmentations from VGAE to improve the effectiveness of contrasting. To cohesively apply latent augmentation for contrastive learning, we propose a novel \textit{Implicit Contrastive Loss} (ICL) with an upper bound for efficient training. Compared to the existing GCL methods based on augmentations with manual perturbations, our approach has the following contributions: 
\begin{itemize}
    \item iGCL exploits latent augmentations that preserves graph topological information, as opposed to classical GCL methods that create augmentations by randomly perturbing edges, nodes or attributes.
    \item The latent augmentations are learned automatically by reconstructing graph structure with a VGAE. Whereas the simple augmentations used in classical GCL methods are designed manually and often require tailored settings for different datasets. 
    \item The proposed \textit{Implicit Contrastive Loss}, unlike classical GCL methods that contrast two augmented views at each time, considers an expected loss w.r.t. all possible latent augmentations that contain graph structural information, and is optimized by an entropy-like upper-bound with a computational-efficient algorithm. This new loss objective is not limited to graph settings but can be applied to broader machine learning fields.
\end{itemize}
\section{Related Work}

\begin{figure*}[t!]
\centering
\includegraphics[width=0.99\textwidth]{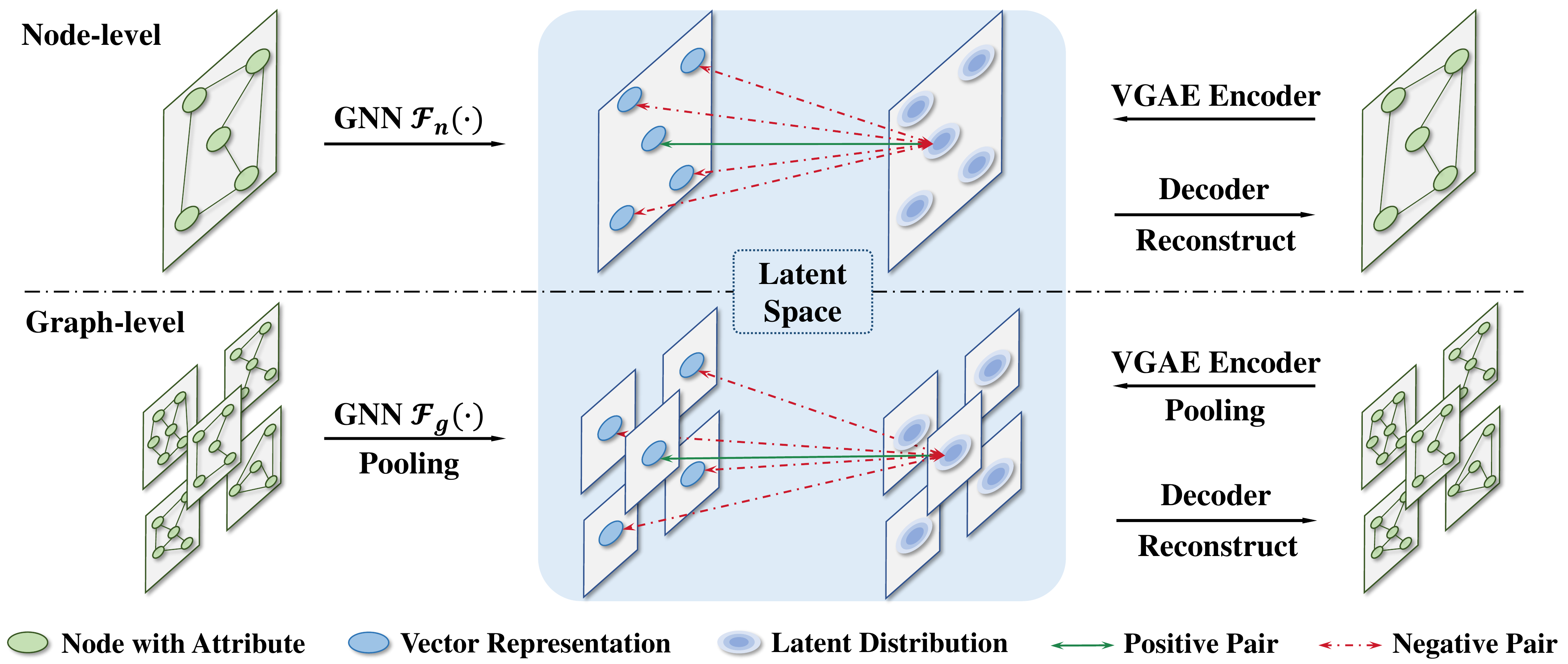} 
\caption{The overall structure of iGCL for node-level task (top) and graph-level task (bottom). For both tasks, we first use a GNN backbone to map the input graph (graphs) into vector representation for each node (graph), and at the same time, we obtain the latent distribution with identical dimension for each node (graph) by training an independent VGAE that reconstructs graph structure. Finally, graph contrastive learning is conducted in the latent space with the proposed \textit{Implicit Contrastive Loss} and its upper bound in Eq.\eqref{eq entropy-like form}, where we regard the vector representation and the latent augmentation for the same node (graph) as positive pair, and others as negative pairs, with details defined in section \ref{sec ICL}.}
\label{fig method}
\end{figure*}

\subsection{Graph Contrastive Learning}
Contrastive learning, an iconic self-supervised learning method that requires no label information for training, has received considerable attention in recent literature \cite{liu2021self}. The general idea is to pull the positive views close in the representation space, and push the negative views apart via maximizing mutual information (MI). For example, Deep InfoMax \cite{hjelm2018learning} contrasts a global feature with other local features through a discriminator to maximize their MI; SimCLR \cite{chen2020simple} pulls augmentations generated from the same input together and push others apart with NT-Xent loss; CPC \cite{oord2018representation} contrasts the context point with other inputs to predict future inputs by optimizing the InfoNCE loss; and MoCo \cite{he2020momentum} further improves InfoNCE by applying momentum to the encoder that generates larger and more consistent dictionary. Currently, most of the graphs contrastive learning methods are extensions of the above works. At node-level, DGI \cite{velivckovic2019deep} mimics Deep InfoMax by maximizing the MI of global representation and local representation; and GRACE \cite{zhu2020deep} treats node representations of augmented views from the same node as positive pairs and others as negative pairs, then optimizes the model in a similar way to SimCLR. At graph-level, InfoGraph \cite{sun2020infograph} extends DGI to graph-level tasks by maximizing the MI between graph-level and patch-level representations; and GraphCL \cite{you2020graph} also adopts NT-Xent from SimCLR that treats augmentations for the same graph as positive pairs and others as negative pairs.

\subsection{Graph Augmentation Methods}
Similar to other fields, augmentation plays an important role in GCL. Common approaches include randomly adding or dropping node and edges \cite{zhu2020deep,you2020graph,xu2021infogcl}, feature shuffling \cite{velivckovic2019deep} and diffusion method \cite{hassani2020contrastive}. However, random augmentations from these methods often change the graph structure and hence lead to meaningless graph views \cite{lee2022augmentation}. To address this issue, latest research begins to use adaptive augmentations such as GCA \cite{zhu2021graph}, or automated augmentation methods such as JOAO \cite{you2021graph} and AutoGCL \cite{yin2022autogcl}. While these works utilize augmentations explicitly, our iGCL takes an implicit way to apply latent augmentations with graph semantic meanings. Unlike existing latent augmentation methods in computer vision \cite{wang2019implicit,li2021metasaug}, the proposed method generates latent augmentations in an unsupervised manner, and are optimized by a novel expected contrastive loss with a computationally efficient upper bound.
\section{Methodology} \label{Sec Methodology}

\subsubsection*{Preliminaries}
Let $\mathcal{G} = (\mathcal{V}, \mathcal{E})$ denote an undirected graph, where $\mathcal{V} = \{v_1, v_2, \cdots, v_N \}$ and $\mathcal{E} \subseteq \mathcal{V} \times \mathcal{V}$ represent the node set and the edge set respectively. ${\bf X} \in \mathbb{R}^{N \times F}$ is the feature matrix for $\mathcal{G}$ with rows $\{ \mathbf{x}^\top_1, \mathbf{x}^\top_2, \cdots, \mathbf{x}^\top_N \}$ ($\mathbf{x}_i$ being a column-vector), 
and ${\bm A} \in \{0,1\}^{N \times N}$ is the adjacency matrix with ${\bm A}_{ij} = 1$ if $(v_i, v_j) \in \mathcal{E}$ and ${\bm A}_{ij} = 0$ otherwise.  Figure~\ref{fig method} demonstrates the overall structure of iGCL. 

The following sections are organized to introduce the methodology of implicit augmentations on graph contrastive learning. In section \ref{sec VGAE}, we demonstrate how we exploit the latent distributions from an independently-trained VGAE as the augmentation source for both node-level and graph-level tasks. In section \ref{sec ICL}, we introduce how we apply the novel \textit{Implicit Contrastive Loss} to train a contrastive learning model by GNN backbones with latent augmentations. In section \ref{sec Discussion}, we visualize the underlying rationale of iGCL and analyze its training object and broader impact.

\subsection{Latent Augmentation with Graph Semantics} \label{sec VGAE}
We aims to train a VGAE that reconstructs graph topological structure, and regard the latent distributions as the source for implicit contrastive learning in the next section. While the proposed algorithm is invariant to the choice of GNNs in VGAE, for simplicity, we adopt a standard 2-layer GCN as the encoder and assume the latent representation at the \textbf{final layer} ${\bm a}_n \in \mathbb{R}^D$ for node $n$ follows a Gaussian distribution $\mathcal{N}({\bm \mu_n}, \, \text{diag} ({\bm \sigma}^2_n)\, )$, with ${\bm \mu}_n$ and ${\bm \sigma}_n$ parameterized by:
\begin{align}
    {{\bf H}_{\bm \mu}^{(\ell)}} & = \text{ReLU}(\Tilde{\bf D}^{-\frac{1}{2}} \Tilde{\bf A} \Tilde{\bf D}^{-\frac{1}{2}} {\bf H}_{\bm \mu}^{(\ell - 1)} {\bf W}_{\bm \mu}^{(\ell)}, \\
    {{\bf H}_{\bm \sigma}^{(\ell)}} & = \text{ReLU}(\Tilde{\bf D}^{-\frac{1}{2}} \Tilde{\bf A} \Tilde{\bf D}^{-\frac{1}{2}} {\bf H}_{\bm \sigma}^{(\ell - 1)} {\bf W}_{\bm \sigma}^{(\ell)} ) \label{encoder},
\end{align}
where $ \Tilde{{\bf D}}_{ii} = \sum_j \Tilde{\bf A}_{ij}$ is the degree matrix of $\Tilde{\bf A}$ with $\Tilde{\bf A} = {\bf A} + {\bf I}$. Both mean and variance encoding networks in Eq.~\eqref{encoder} share the same weights in the first layer (i.e. ${\bf W}_{\bm \mu}^{(0)} = {\bf W}_{\bm \sigma}^{(0)}$) and use $\text{ReLU}(t) = \max(0,\,t)$ as the activation function. Note ${\bm \mu}_n^{(\ell)}$ and $\log {\bm \sigma}_n^{(\ell)}$ are both row-vectors in ${{\bf H}_{\bm \mu}^{(\ell)}}$ and ${{\bf H}_{\bm \sigma}^{(\ell)}}$ at the $\ell$-th layer respectively, except for $\ell = 0$ where ${\bf H}_{\bm \sigma}^{(0)} = {\bf H}_{\bm \mu}^{(0)} = {\bf X}$.

The model is then optimized by maximizing the evidence lower bound $\mathcal{L}_{VGAE}$ from variational inference:
\begin{equation}
    \mathcal{L}_{VGAE} = \mathbb{E}_{q({\bf H}|{\bf X}, {\bf A})}[\log p({\bf A}|{\bf H})] - \text{KL}[q({\bf H}|{\bf X},{\bf A})||p({\bf H})] \label{eq VGAE}
\end{equation}
where the first term is the likelihood for reconstruction by an inner-product decoder $p({\bf A}|{\bf H}) = \prod_{i=1}^N \prod_{j=1}^{N}\sigma({\bm a}_i^\top{\bm a}_j)$ ($\sigma(\cdot)$ is the logistic sigmoid function). The second term regularizes the Kullback-Leibler divergence between the variational distribution $q({\bf H}|{\bf X},{\bf A}) = \prod_{n=1}^{N} q({\bm a}_n|{\bf X},{\bf A})$ (parameterized by the encoder networks) and a chosen prior $p({\bf H}) = \prod_{n=1}^N p ({\bm a}_n) = \prod_{n=1}^N \mathcal{N}({\bm a}_n|{\bf 0}, {\bf I})$ (we adopt the standard Gaussian prior from the original paper). Upon convergence, we consider latent augmentations for node-level tasks and graph-level tasks separately below.

\subsubsection{Latent Augmentation for Node-level Task}
We define a latent augmentation ${\tilde {\bm a}}_n \in \mathbb{R}^D$ for node $v_n$ as a sample from the learned latent distribution:
\begin{equation}
    {\tilde {\bm a}}_n \sim \mathcal{N} ({\bm \mu}_n, {\bm \Sigma}_n), \label{eq node latent distribution}
\end{equation} 
where ${\bm \Sigma}_n$ is a diagonal matrix with elements in ${\bm \sigma}^2_n$. These augmentations contains topological information of the original graph: they can reconstruct the graph by a similarity matrix $\Bar{{\bf A}}$ of the original adjacency matrix $\bf A$ by computing $\Bar{{\bf A}}_{ij} = \sigma ({\tilde {\bm a}}_i^\top {\tilde {\bm a}}_j)$.

\subsubsection{Latent Augmentation for Graph-level Task}
Given a set of $G$ graphs ${\{ \mathcal{G}_g \}}_{g=1}^G$, we train a single VGAE on multiple graphs by reconstructing their graph structures, with training objective in Eq. \eqref{eq VGAE} replaced by:   
\begin{equation}
    \mathcal{L}_{total} = \frac{1}{G}\sum_{g=1}^{G} \mathcal{L}_{VGAE}^{(g)}.\label{eq VGAE graph}
\end{equation}
Then, we apply graph pooling in the embedding space to get an global representation ${\bm r}_g$ for graph $\mathcal{G}_g$, which is also assumed to follow a Gaussian distribution $\mathcal{N} ({\bm \mu}_g, \, \text{diag}({\bm \sigma}_g^2))$, with ${\bm \mu}_g$ and ${\bm \sigma}_g$ parameterized by a mean-aggregator used in \cite{garnelo2018conditional,garnelo2018neural}:
\begin{equation}
    {\bm \mu }_g = \frac{1}{N} \sum_{n=1}^N {\bm \mu}_n, \; \log {\bm \sigma }_g = \frac{1}{N} \sum_{n=1}^N \log {\bm \sigma_n},\label{eq graph mu}
\end{equation}
where $N$ is the number of nodes on graph $\mathcal{G}_g$ and it can vary among different graphs. Similar to Eq. \eqref{eq node latent distribution} for node-level augmentation, we can now define a latent augmentation $\tilde{\bm r}_g \in \mathbb{R}^D$ for graph $\mathcal{G}_g$ as a sample from the latent distribution.

\subsection{Implicit Graph Contrastive Learning}
\begin{figure}[t!]
\centering\includegraphics[width=0.99\columnwidth]{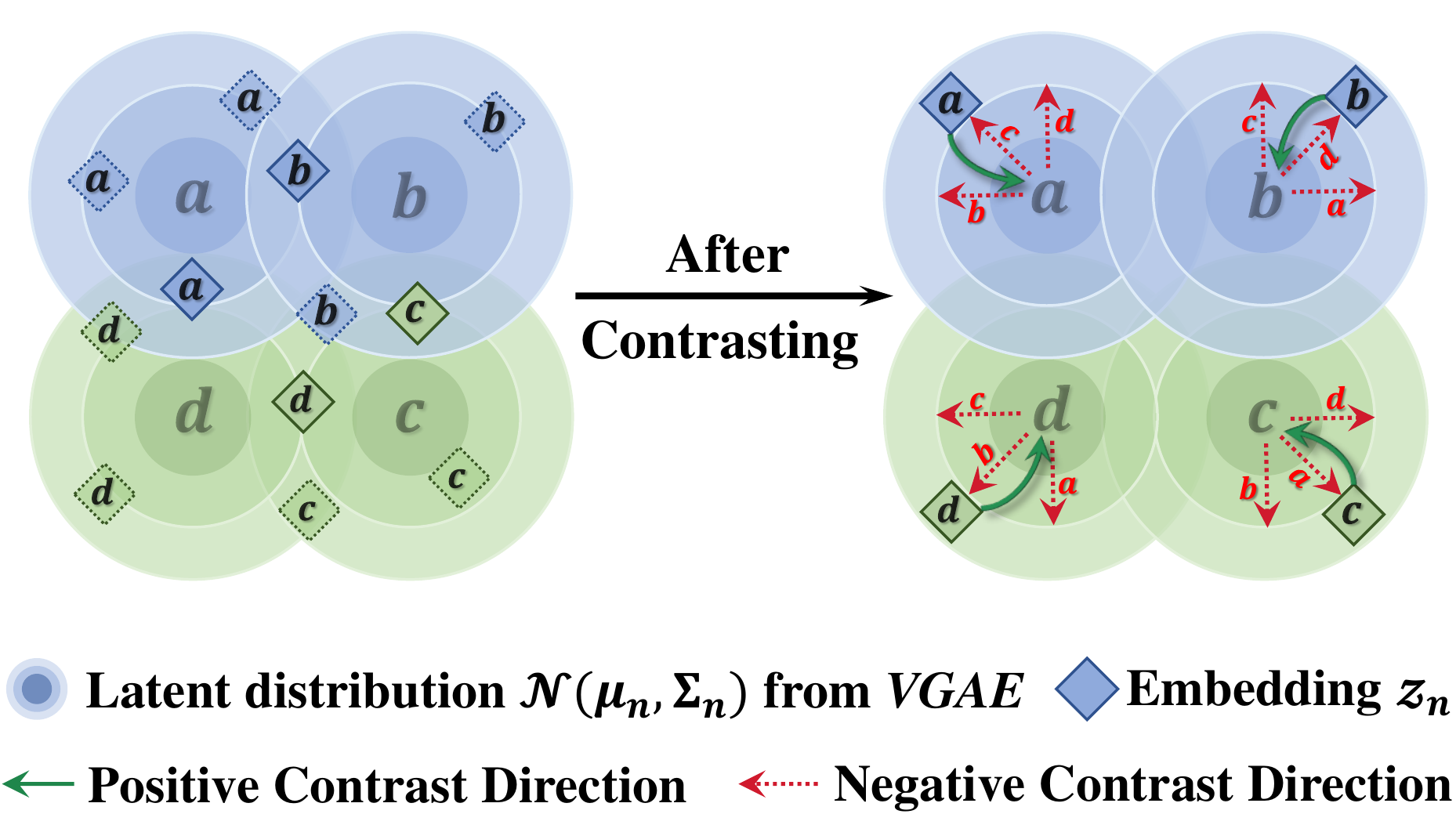} 
\caption{Illustration for the underlying behaviour of iGCL in Eq.~\eqref{eq entropy-like form}. Suppose we generate latent distributions for nodes (graphs) $a, b, c, d$ by VGAE and sample from them, the resulting embeddings (in dashed squares) could be hard to distinguish from their source distributions due to overlapping problem. However, after contrasting, embedding of node $v_n$ (graph $\mathcal{G}_n$) is pushed away from the other nodes (graphs) distributions but also pulled close to its own distribution. Therefore, these new embeddings become more separated, and meanwhile preserve the graph structural information.}
\label{fig interpretation}
\end{figure}

After obtaining the latent augmentations, at both node-level and graph-level, we train a GNN backbone by the proposed \textit{Implicit Contrastive Loss} (ICL) below, with its underlying behaviour illustrated in Figure~\ref{fig interpretation}.
\subsubsection{GNN Backbone}
We consider two types of GNN backbones at node-level and graph-level: $\mathcal{F}_n({\bf X},{\bf A}):\mathbb{R}^{N \times F}\times \mathbb{R}^{N \times N} \rightarrow \mathbb{R}^{N \times D}$ that maps the input graph into $N$ vector representations for node-level tasks, and $\mathcal{F}_g({\bf X},{\bf A}):\mathbb{R}^{N \times F}\times \mathbb{R}^{N \times N} \rightarrow \mathbb{R}^{D}$ that maps the input graph into a global vector representation for graph-level tasks. Following the common choices in GCL literature, we adopt GCN or GAT as $\mathcal{F}_n(\cdot)$ and GIN \cite{xu2019powerful} as $\mathcal{F}_g(\cdot)$. A non-linear projection head $\mathcal{P}(\cdot)$ (2-layer MLP) is also applied after the backbone to improve the expressive power \cite{chen2020simple}. Since both node-level and graph-level tasks aim to generate a vector representation for each node or for each graph respectively, for simplicity purpose, we will slightly abuse the notation ${\bm z}_n$ to represent the vector representation for node $v_n$ in node-level tasks, or for graph ${\mathcal{G}}_n$ in graph-level tasks for the rest of the paper. 

\begin{algorithm}[t]
\caption{iGCL for Graph-level Task} \label{alg algorithm graph}
\textbf{Input}: A set of graphs $\{\mathcal{G}_n\}_{n=1}^{N}$; Graph Auto-Encoder $VGAE(\cdot)$; GNN backbone $\mathcal{F}(\cdot)$ and projection head $\mathcal{P}(\cdot)$.\\
\textbf{Parameter}: $\Phi$ in $VGAE(\cdot)$; $\Theta$ in $\mathcal{F}(\cdot)$ and $\mathcal{P}(\cdot)$.\\
\vspace{-1.2em}
\begin{algorithmic}[1] 
\For{$epoch = 1,\cdots,T$}
    \For{$batch$ \textbf{in} $\{\mathcal{G}_n\}_{n=1}^{N}$}
    \State Generate ${\bm \mu}_n, {\bm \sigma}_n$ for $\mathcal{G}_n$ from $VGAE$ Eq.~\eqref{eq graph mu}.
    \For{$k = 1,\cdots, K$}
        \State Update $\Phi$ by $\nabla \mathcal{L}_{total}$ in Eq.~\eqref{eq VGAE graph}.
    \EndFor
    \State Generate ${\bm z}_n$ for graph $\mathcal{G}_n$ from $\mathcal{F}(\cdot)$ and $\mathcal{P}(\cdot)$.
    \State Update $\Theta$ by $\nabla \mathcal{L}_{upper}$ in Eq.~\eqref{eq entropy-like form}.
    \EndFor
\EndFor
\State \textbf{return} Embedding $\bf Z$ from the optimized $\mathcal{F}(\cdot)$.
\end{algorithmic}
\end{algorithm}

\subsubsection{Implicit Contrastive Loss} \label{sec ICL}
Following the \textit{InfoNCE} \cite{oord2018representation} objective in contrastive learning, we sample $M$ times from the latent distribution $\mathcal{N} ({\bm \mu}_n, {\bm \Sigma}_n)$ in Eq.\eqref{eq node latent distribution}, and denote ${\bm a}_n^m$ as the $m$-th latent augmentation for node $v_n$ in node-level tasks, or for graph $\mathcal{G}_n$ in graph-level tasks. As demonstrated in Figure~\ref{fig method}, for each augmentation sample, we regard (${\bm a}_n^m$, ${\bm z}_n$) as the positive pair, and $\{({\bm a}_n^m ,{\bm z}_{n'})\}_{[n' \neq n]}$ as the negative pairs. Then the contrastive loss $\mathcal{L}_{CL}$ for $M$ augmentation samples is given by:
\begin{equation}
    \mathcal{L}_{CL} = \frac{1}{N}\sum_{n=1}^{N}\frac{1}{M}\sum_{m=1}^{M} -\log \frac{\exp({\bm z}_n ^\top {\bm a}_n^{m} /\tau)}{\sum_{n'=1}^{N} \exp({\bm z}_{n'}^\top {\bm a}_n^m/\tau)}, \label{eq CL}
\end{equation}
where we use inner-product scaled by a temperature hyper-parameter $\tau$ as the similarity estimator. 

Although a large $M$ will improve the effectiveness of $\mathcal{L}_{CL}$, the algorithm becomes highly computational inefficient during optimization as the augmentation set is enlarged by $M$ times with a complexity of $\mathcal{O}(MN^2D^2)$. To solve this problem, in the following section, we will investigate the situation when $M$ grows to infinity, and propose an entropy-like upper bound that can be easily computed, hence leading to a more efficient algorithm.

\begin{algorithm}[t!]
\caption{iGCL for Node-level Task} \label{alg algorithm node}
\textbf{Input}: Graph $\mathcal{G}$ with $N$ nodes; Graph Auto-Encoder $VGAE(\cdot)$; GNN backbone $\mathcal{F}(\cdot)$ and projection head $\mathcal{P}(\cdot)$.\\
\textbf{Parameter}: $\Phi$ in $VGAE(\cdot)$; $\Theta$ in $\mathcal{F}(\cdot)$ and $\mathcal{P}(\cdot)$.\\
\vspace{-1.2em}
\begin{algorithmic}[1] 
\For{$epoch = 1,\cdots,T$}
    \State Generate ${\bm \mu}_n, {\bm \sigma}_n$ for $v_n$ from $VGAE$.
    \For{$k = 1,\cdots, K$}
        \State Update $\Phi$ by $\nabla \mathcal{L}_{VGAE}$ in Eq.~\eqref{eq VGAE}.
    \EndFor
    \State Generate ${\bm z}_n$ for node $v_n$ from $\mathcal{F}(\cdot)$ and $\mathcal{P}(\cdot)$.
    \State Randomly sample a batch of size $b$ from $\{z_n\}_{n=1}^N$
    \State Compute $\mathcal{L}_{upper}$ in Eq.~\eqref{eq entropy-like form} within the batch.
    \State Update $\Theta$ by $\nabla \mathcal{L}_{upper}$.
\EndFor
\State \textbf{return} Embedding $\bf Z$ from the optimized $\mathcal{F}(\cdot)$.
\end{algorithmic}
\end{algorithm}

\begin{table*}[t!]
\centering
\setlength{\tabcolsep}{9 pt} 
\renewcommand{\arraystretch}{1} 
\resizebox{0.99\textwidth}{!}{
\begin{tabular}{c | l c c c c c c }
    \hline\hline
    \textbf{Graph} & \textbf{Dataset}  & \textbf{MUTAG} & \textbf{NCI1} & \textbf{PROTEINS} & \textbf{COLLAB} & \textbf{IMDB-B} & \textbf{IMDB-M}\\
    \hline\hline
    \multirow{5}{*}{\textbf{Statistics}} 
    & Type         & Molecule & Molecule & Molecule & Social & Social & Social \\
    & \# Graphs    & 188 & 4,110 & 1,113 & 5,000 & 1,000 & 1,500 \\
    & \# Classes   & 2 & 2 & 2 & 3 & 2 & 3\\
    & Avg. Nodes   & 17.9 & 29.9 & 39.1 & 74.5& 19.8 & 13.0 \\
    & Avg. Edges   & 19.8 & 32.3 & 72.8 & 2,457.8 & 96.5 & 65.9\\
    \hline
    \multirow{6}{*}{\textbf{Settings}}
    & Projection  & Skip & Linear & Linear & Linear & Linear & Skip \\
    & num\_epochs & 100 & 100 & 20 & 20 & 20 & 100\\
    & num\_layers & 5 & 5 & 2 & 8 & 2 & 8\\
    & emb\_size   & 256 & 256 & 256 & 256 & 512 & 128 \\
    & batch\_size & 16 & 32 & 64 & 128 & 16 & 64\\
    & $lr$        & $5\times10^{-4}$ & $10^{-4}$ & $10^{-4}$ & $5\times10^{-4}$ & $10^{-4}$ & $5 \times 10^{-4}$ \\
    & $\ell_2$    & $5\times 10^{-3}$ & $5\times 10^{-3}$ & $10^{-2}$ & $10^{-2}$ & $5\times10^{-3}$ & $10^{-2}$ \\
    & $\tau$      & 0.01 & 3.54 & 5 & 1.98 & 5.0 & 10.0 \\
    \hline\hline
\end{tabular}}
\caption{Dataset statistics and hyper-parameter settings for Graph-level tasks.}
\label{tab Graph stats}
\end{table*}

\subsubsection*{Upper Bound} 
As $M \rightarrow \infty$, we are in fact discussing the expectation of the above contrastive loss for all possible latent augmentations, which is defined as the \textit{Implicit Contrastive Loss} $\mathcal{L}_{ICL} := \lim \limits_{M \to +\infty} \mathcal{L}_{CL}$ as follows:
\begin{align}
    \mathcal{L}_{ICL} & = \frac{1}{N}\sum_{n=1}^{N} \mathbb{E}_{\tilde {\bm a}_n} \Big[ -\log \frac{\exp({\bm z}_n ^\top {\tilde {\bm a}_n} /\tau)}{\sum_{n'=1}^{N} \exp({\bm z}_{n'}^\top{\tilde {\bm a}}_n/\tau)} \Big]\\
    & = \frac{1}{N}\sum_{n=1}^{N} \mathbb{E}_{\tilde {\bm a}_n} \Big[ \log \sum_{n'=1}^{N} \exp \big( ({\bm z}_{n'} - {\bm z}_n)^\top \, {\tilde {\bm a}}_n/\tau \big) \Big].
\end{align}
Given the Jensen's inequality $\mathbb{E}[\log X] \leq \log \mathbb{E}[X]$,  we can find an upper bound $\mathcal{L}_{upper}$ for $\mathcal{L}_{ICL}$  by swapping the log and expectation, that is, $\mathcal{L}_{upper} \geq \mathcal{L}_{ICL}$ with:
\begin{align}
    \mathcal{L}_{upper} & = \frac{1}{N}\sum_{n=1}^{N}\log \mathbb{E}_{\tilde {\bm a}_n} \Big[\sum_{n'=1}^{N} \exp \big( ({\bm z}_{n'} - {\bm z}_n)^\top \, {\tilde {\bm a}}_n/\tau \big) \Big]. \label{eq Jensen}
\end{align}
Since ${\bm \delta}_{n'n} := {\bm z}_{n'} - {\bm z}_n$ is a constant inside the expectation in Eq.\eqref{eq Jensen} and $\tilde{\bm a}_n$ follows $\mathcal{N} ({\bm \mu}_n, {\bm \Sigma}_n)$, we can rewrite $\mathcal{L}_{upper}$ with the moment generating function:
\begin{equation}
    \mathbb{E}[\exp(tX)] = \exp(t \mu + \frac{1}{2}\sigma^2t^2), \; X \sim \mathcal{N}(\mu,\sigma^2),
\end{equation}
which transforms the expectation form of $\mathcal{L}_{upper}$ into:
\begin{equation}
    \frac{1}{N}\sum_{n=1}^{N}\log \sum_{n'=1}^N \exp \big( {\bm \delta}_{n'n}^\top \, {\bm \mu}_n/\tau + {\bm \delta}_{n'n}^\top \, {\bm \Sigma}_n \, {\bm \delta}_{n'n}/2\tau^2 \big). \label{eq upper bound}
\end{equation}
Finally, we restore Eq.\eqref{eq upper bound} to an entropy-like equivalent:
\begin{equation}
    \frac{1}{N}\sum_{n=1}^{N} - \log \frac{\exp({\bm z}_n^\top {\bm \mu}_n/\tau)}{\sum_{n'=1}^N \exp \big({\bm z}_{n'}^\top{\bm \mu}_n/\tau + {\bm \sigma}_n \odot {\bm \delta}_{n'n}^\top \,{\bm \delta}_{n'n}/2\tau^2 \big)} \label{eq entropy-like form}
\end{equation}
where we further simplify the quadratic term ${\bm \delta}_{n'n}^\top \, {\bm \Sigma}_n \, {\bm \delta}_{n'n}$ into an element-wise operation ${\bm \sigma}_n \odot {\bm \delta}_{n'n}^\top \,{\bm \delta}_{n'n}$, since ${\bm \Sigma}_n$ is a diagonal matrix with diagonal entries ${\bm \sigma}_n$. Compared to $\mathcal{L}_{CL}$ in Eq.~\eqref{eq CL}, the proposed upper bound in Eq.~\eqref{eq entropy-like form} is more efficient with a complexity of $\mathcal{O}(N^2D^2)$. We demonstrate the learning algorithms for iGCL at graph-level and node-level in Algorithm \ref{alg algorithm graph} and Algorithm \ref{alg algorithm node}.

\subsection{Interpretation of iGCL and Broader Impact} \label{sec Discussion}
We present a visual explanation shown in Figure~\ref{fig interpretation} to explain the underlying behaviour of applying implicit contrastive loss ICL in Eq.~\eqref{eq entropy-like form}. The objective of iGCL is to contrasts on the latent distributions from VGAE, where embedding ${\bm z}_n$ for node $v_n$ (graph $\mathcal{G}_n$) are pushed away from the means of the other nodes' latent distributions, and pulled close to the mean of its own latent distribution, with temperature $\tau$ controlling the contrastive scale such that it won't be excessively far away for reconstruction. In this way, ICL enables a GNN backbone $\mathcal{F}(\cdot)$ to separate the embeddings of different nodes (or graphs) and maintain graph semantic information at the same time. Moreover, compared to existing GCL methods such as GraphCL (from SimCLR) and InfoGraph (from Deep InfoMax) that only contrast two augmented views at each training epoch, iGCL contrast with the distributions of all possible latent augmentations at each iteration but requires the same level of complexity, credited to the computational-efficient upper bound.

The proposed \textit{Implicit Contrastive Learning} approach does not only limit to graph settings and can be extended to other machine learning fields. The general idea is to first train a VAE that can generate high-level latent distributions with semantic meanings, then regard them as the source for latent augmentations and conduct implicit contrastive learning with the proposed upper bound for ICL in Eq.~\eqref{eq entropy-like form}. To the best of our knowledge, no similar approach has been explored in the existing literature.

\linespread{1.2}
\begin{table*}[t!]
\centering
\setlength{\tabcolsep}{8.5 pt} 
\renewcommand{\arraystretch}{1} 
\resizebox{0.99\textwidth}{!}{
\begin{tabular}{c | l c c c c c c}
    \hline\hline
    \textbf{Input} & \textbf{Method} & \textbf{MUTAG} & \textbf{NCI1} & \textbf{PROTEINS} & \textbf{COLLAB} & \textbf{IMDB-B} & \textbf{IMDB-M} \\
    \hline\hline
    \multirow{2}{*}{$\bf A, X, Y$} 
     & GCN  & 85.6 $\pm$ 5.8 & 80.2 $\pm$ 2.0 & 74.9 $\pm$ 3.3 & 79.0 $\pm$ 1.8 & 70.4 $\pm$ 3.4 & 51.9 $\pm$ 3.8 \\
     & GIN  & 89.4 $\pm$ 5.6 & 82.7 $\pm$ 1.7 & 76.2 $\pm$ 2.8 & 80.2 $\pm$ 1.9 & 75.1 $\pm$ 5.1 & 52.3 $\pm$ 2.8 \\
    \hline
    \multirow{8}{*}{$\bf A, X$}
     & node2vec & 72.6 $\pm$ 10.2& 54.9 $\pm$ 1.6 & 57.5 $\pm$ 3.6 & 56.1 $\pm$ 0.2 & 50.2 $\pm$ 0.9 & 36.0 $\pm$ 0.7\\
     & sub2vec  & 61.1 $\pm$ 15.9& 52.8 $\pm$ 1.5 & 53.0 $\pm$ 5.6 &       -        & 55.3 $\pm$ 1.5 & 36.7 $\pm$ 0.8 \\
     & graph2vec & 83.2 $\pm$ 9.3 & 73.2 $\pm$ 1.8 & 73.3 $\pm$ 2.1 &       -        & 71.1 $\pm$ 0.5 & \textbf{50.4} $\pm$ \textbf{0.9} \\
     & InfoGraph & \textcolor{blue}{89.0} $\pm$ \textcolor{blue}{1.1} & 76.2 $\pm$ 1.0 & 74.4 $\pm$ 0.3 & 70.7 $\pm$ 1.1 & \textcolor{blue}{73.0} $\pm$ \textcolor{blue}{0.9} & 49.7 $\pm$ 0.5 \\
     & GraphCL  & 86.8 $\pm$ 1.3 & 77.9 $\pm$ 0.4 & 74.4 $\pm$ 0.5 & \textcolor{blue}{71.4} $\pm$ \textcolor{blue}{1.1} & 71.1 $\pm$ 0.4 & 49.2 $\pm$ 0.6 \\
     & JOAO    & 87.4 $\pm$ 1.0 & 78.1 $\pm$ 0.5 & 74.6 $\pm$ 0.4 & 69.5 $\pm$ 0.4 & 70.2 $\pm$ 3.1 & - \\
     & AutoGCL & 88.6 $\pm$ 1.1 & \textcolor{blue}{82.0} $\pm$ \textcolor{blue}{0.3} & \textbf{75.8} $\pm$ \textbf{0.4}  & 70.1 $\pm$ 0.7 & \textbf{73.3} $\pm$ \textbf{0.4} & - \\
     & \textbf{iGCL} (ours) & \textbf{89.8} $\pm$ \textbf{1.2} & \textbf{82.7} $\pm$ \textbf{0.4} & \textcolor{blue}{74.8} $\pm$ \textcolor{blue}{0.5} & \textbf{72.0} $\pm$ \textbf{0.8} & 72.6 $\pm$ 0.6& \textcolor{blue}{49.8} $\pm$ \textcolor{blue}{0.6}\\
    \hline\hline
\end{tabular}}
\caption{Graph Classification Results. Numbers highlighted in \textbf{bold} and \textcolor{blue}{blue} represent the \textbf{best} and the \textcolor{blue}{second} best results respectively.}
\label{tab Graph Classification}
\end{table*}

\vspace{-0.3 cm}
\section{Experiment} \label{Sec Experiment}
In this section, we empirically evaluate our proposed method on graph-level and node-level tasks, and compare the performance with other state-of-the-art GNN methods. An ablation study is also conducted in the end to discuss the effectiveness of different modules in our model. Code for reproducing the experiment results can be found in the supplementary materials.

\subsection{Graph-level Tasks}
\subsubsection{Settings}
For graph-level tasks, we test our model by classification task on three biochemical molecule networks: MUTAG \cite{debnath1991structure}, NCI1\cite{wale2008comparison} and PROTEINS \cite{borgwardt2005protein}; and three social networks: COLLAB \cite{leskovec2005graphs}, IMDB-B and IMDB-M \cite{yanardag2015deep}, where detailed dataset statistics are summarized in Table~\ref{tab Graph stats}. The selected baselines for comparison are three classic unsupervised methods: node2vec \cite{grover2016node2vec}, sub2vec \cite{adhikari2018sub2vec} and graph2vec \cite{narayanan2017graph2vec}; two common GCL baselines: InfoGraph \cite{sun2020infograph} and GraphCL\cite{you2020graph}; and two latest GCL methods with automated augmentations: JOAO \cite{you2021graph} and AutoGCL \cite{yin2022autogcl}. The experiment results for these baselines are adopted from their original papers or reproduced by the official codes released from their authors.

\subsubsection{Graph Classification}

We closely follow the common approach in GCL for graph classification \cite{sun2020infograph,you2020graph}: the GNN backbone is chosen as GIN \cite{xu2019powerful} with layers from \{2,5,8\} and embedding size from \{128, 256, 512\}. After contrastive learning, we use a SVM as the classifier with hyper-parameter $C$ chosen from $\{ 10^{-3}, 10^{-2}, \cdots, 10^{2}, 10^{3}\}$ and report the 10-fold cross validation accuracy as the final result. The experiments are repeated 10 times for graph classification, with evaluation accuracy reported by mean and standard deviation. Detailed hyper-parameter settings for reproduction are summarized in Table~\ref{tab Graph stats}.

The results in Table~\ref{tab Graph Classification} indicate that iGCL outperforms classic GCL baselines InfoGraph and GraphCL on five datasets and achieves comparable result on IMDB-B. Compared with JOAO and AutoGCL that use automated augmentation method in GCL, iGCL also shows the best performance on MUTAG, NCI1 and COLLAB, and ranks the second best on PROTEINS and IMDB-M.

\subsection{Node-level Tasks}
\subsubsection{Settings}
In this section, we validate iGCL on node-level classification and clustering tasks. The datasets used for evaluation are three citation networks: Cora, Citeseer and PubMed \cite{sen2008collective}; and two Amazon sales dataset: Computers and Photo \cite{mcauley2015image}, where the statistics are summarized in Table~\ref{tab Node stats}. For node classification tasks, We compare our model against two GCL methods with random augmentations: DGI \cite{velivckovic2019deep} and GRACE \cite{zhu2020deep}; a non-augmentation method GMI \cite{peng2020graph}; and an adaptive augmentation method GCA \cite{zhu2021graph}. For node clustering, we compare iGCL to three classical clustering methods on graphs: K-means, Spectral Clustering~\cite{tang2011leveraging} and DeepWalk~\cite{perozzi2014deepwalk}; three Graph Auto-Encoder models: VGAE~\cite{kipf2016variational}, ARGA~\cite{pan2018adversarially} and GALA~\cite{park2019symmetric}, plus a contrastive method MVGRL~\cite{hassani2020contrastive}. The results are retained from original papers or reproduced by the official codes from their authors.

\subsubsection{Node Classification}
\begin{table*}[t]
\centering
\setlength{\tabcolsep}{10 pt} 
\renewcommand{\arraystretch}{0.85} 
\resizebox{0.99\textwidth}{!}{
\begin{tabular}{c | l c c c c c }
    \hline\hline
    \textbf{Node} & \textbf{Dataset}  & \textbf{Cora} & \textbf{Citeseer} & \textbf{PubMed} & \textbf{Photo} & \textbf{Computers}\\
    \hline\hline
    \multirow{5}{*}{\textbf{Statistics}} 
    & Type         & Citation & Citation & Citation & Amazon Sales & Amazon Sales \\
    & \# Features  & 1,433 & 3,703 & 500 & 745 & 767\\
    & \# Classes   & 7 & 6 & 3 & 8 & 10\\
    & \# Nodes     & 2,708 & 3,327 & 19,717 & 7,650 & 13,752\\
    & \# Edges     & 5,429 & 4,732 & 44,338 & 119,081 & 245,861\\
    \hline
    \multirow{6}{*}{\textbf{Settings}}
    & Backbone  & GCN & GAT & GCN & GAT & GAT \\
    & num\_epochs & 300 & 300 & 1000 & 600 & 600 \\
    & emb\_size   & 256 & 512 & 128 & 256 & 256 \\
    & dropout   & 0.5 & 0.5 & 0.5 & 0.5 & 0.5 \\
    & $lr$ & $10^{-4}$ & $5\times10^{-4}$ & $10^{-4}$ & $5\times10^{-3}$ & $10^{-3}$ \\
    & $lr$ step & - & 0.85 & - & 0.85 & 0.85 \\
    & $\ell_2$ & $5\times 10^{-3}$ & $10^{-2}$ & $10^{-4}$ & $5\times 10^{-3}$ & $5\times 10^{-3}$ \\
    & $\tau$ & 1.0 & 5.0 & 4.7 & 1.0 & 1.0 \\
    \hline\hline
\end{tabular}}
\caption{Dataset statistics and hyper-parameter settings for Node-level task.}
\label{tab Node stats}
\end{table*}

\begin{table*}[t!]
\centering
\setlength{\tabcolsep}{12 pt } 
\renewcommand{\arraystretch}{1} 
\resizebox{0.99\textwidth}{!}{
\begin{tabular}{c | l c c c c c}
    \hline\hline
    \textbf{Input} & \textbf{Method} & \textbf{Cora} & \textbf{Citeseer} & \textbf{PubMed} & \textbf{Photo} & \textbf{Computers}  \\
    \hline\hline
    \multirow{3}{*}{$\bf A, X, Y$} 
    & GCN   & 81.5 $\pm$ 0.2 & 70.3 $\pm$ 0.4 & 79.0 $\pm$ 0.5 & 92.4 $\pm$ 0.2 & 86.5 $\pm$ 0.5 \\
    & GraphSAGE & 83.0 $\pm$ 0.7 & 72.5 $\pm$ 0.7 & 79.0 $\pm$ 0.3 & 92.6 $\pm$ 0.4 & 86.9 $\pm$ 0.3 \\
    & GAT  & 83.0 $\pm$ 0.7 & 72.5 $\pm$ 0.7 & 79.0 $\pm$ 0.3 & 92.6 $\pm$ 0.4 & 86.9 $\pm$ 0.3 \\
    \hline
    \multirow{7}{*}{$\bf A, X$}
    & GAE  & 74.9 $\pm$ 0.4 & 65.6 $\pm$ 0.5 & 74.2 $\pm$ 0.3 & 91.6 $\pm$ 0.1 & 85.3 $\pm$ 0.2 \\
    & VGAE & 76.3 $\pm$ 0.2 & 66.8 $\pm$ 0.2 & 75.8 $\pm$ 0.4 & 92.2 $\pm$ 0.1 & 86.4 $\pm$ 0.2 \\
    & DGI  & 82.3 $\pm$ 0.6 & 71.8 $\pm$ 0.7 & 76.8 $\pm$ 0.6 & 91.6 $\pm$ 0.2 & 84.0 $\pm$ 0.5 \\
    & GMI  & 83.0 $\pm$ 0.3 & \textbf{73.0} $\pm$ \textbf{0.3} & 80.1 $\pm$ 0.2 & 90.7 $\pm$ 0.2 & 82.2 $\pm$ 0.3 \\
    & GRACE & \textcolor{blue}{83.3} $\pm$ \textcolor{blue}{0.4} & 72.1 $\pm$ 0.5 & 80.6 $\pm$ 0.4 & 91.5 $\pm$ 0.4 & 87.1 $\pm$ 0.3 \\
    & GCA  & 83.1 $\pm$ 0.3 & \textcolor{blue}{72.3} $\pm$ \textcolor{blue}{0.5} & \textbf{80.9} $\pm$ \textbf{0.3} & \textbf{92.5} $\pm$ \textbf{0.2} & \textcolor{blue}{87.8} $\pm$ \textcolor{blue}{0.3} \\
    & \textbf{iGCL} (ours) & \textbf{83.6} $\pm$ \textbf{0.6} & 72.2 $\pm$ 0.6 & \textcolor{blue}{80.8} $\pm$ \textcolor{blue}{0.6} & \textbf{92.5} $\pm$ \textcolor{blue}{0.5} & \textbf{88.1} $\pm$ \textbf{0.3}\\
    \hline\hline
\end{tabular}}
\caption{Node Classification Results. Numbers highlighted in \textbf{bold} and \textcolor{blue}{blue} represent the \textbf{best} and the \textcolor{blue}{second} best results respectively.}
\label{tab Node Classification}
\end{table*} 

\begin{table*}[t!]
\centering
\setlength{\tabcolsep}{10 pt } 
\renewcommand{\arraystretch}{1} 
\resizebox{0.99\textwidth}{!}{
\begin{tabular}{c | l c c c c c c c c c}
    \hline\hline
    \multirow{2}{*}{\textbf{Input}} & \multirow{2}{*}{\textbf{Method}} & \multicolumn{3}{c}{\textbf{Cora}} & \multicolumn{3}{c}{\textbf{Citeseer}} & \multicolumn{3}{c}{\textbf{PubMed}}\\ \cline{3-11}
    & & Acc & NMI & ARI & Acc & NMI & ARI & Acc & NMI & ARI \\
    \hline\hline
    \multirow{3}{*}{$\bf A$} 
    & K-means & 0.492 & 0.321 & 0.230 & 0.540 & 0.305 & 0.279 & 0.398 & 0.001 & 0.002 \\
    & SC & 0.367 & 0.127 & 0.031 & 0.239 & 0.056 & 0.010 & 0.403 & 0.042 & 0.002 \\
    & DeepWalk & 0.484 & 0.327 & 0.243 & 0.337 & 0.088 & 0.092 & 0.684 & 0.279 & 0.299 \\
    \hline
    \multirow{7}{*}{$\bf A, X$}
    & GAE   & 0.596 & 0.429 & 0.347 & 0.408 & 0.176 & 0.124 & 0.672 & 0.277 & 0.279 \\
    & VGAE  & 0.443 & 0.239 & 0.175 & 0.344 & 0.156 & 0.093 & 0.630 & 0.229 & 0.213 \\
    & ARGA  & 0.640 & 0.449 & 0.352 & 0.573 & 0.350 & 0.341 & 0.668 & 0.305 & 0.295 \\
    & ARVGA & 0.638 & 0.450 & 0.374 & 0.544 & 0.261 & 0.245 & 0.690 & 0.290 & 0.306 \\
    & GALA  & \textbf{0.746} & \textbf{0.577} & \textcolor{blue}{0.532} & \textcolor{blue}{0.693} & \textbf{0.441} & 0.446 & \textcolor{blue}{0.694} & \textcolor{blue}{0.327} & \textcolor{blue}{0.321} \\
    & MVGRL & 0.732 & \textcolor{blue}{0.569} & 0.522 & 0.691 & \textcolor{blue}{0.437} & \textcolor{blue}{0.449} & 0.687 & 0.319 & 0.315 \\
    & \textbf{iGCL} & \textcolor{blue}{0.744} & 0.566 & \textbf{0.539} & \textbf{0.695} & 0.435 & \textbf{0.452} & \textbf{0.697} & \textbf{0.334} & \textbf{0.329}\\
    \hline\hline
\end{tabular}}
\caption{Node Clustering Results. Numbers highlighted in \textbf{bold} and \textcolor{blue}{blue} represent the \textbf{best} and the \textcolor{blue}{second} best results respectively.}
\label{tab Node Clustering}
\end{table*} 

For node classification task, we adopt the standard dataset splits for citation networks \cite{kipf2017semi} and Amazon sales \cite{zhu2021graph}. We follow the common search strategy \cite{zhu2020deep,zhu2021graph} in GCL for our model configurations: the GNN backbone is selected as GCN \cite{kipf2017semi} or GAT \cite{velivckovic2018graph} with layers from {1, 2, 3} and embedding sizes from {128, 256, 512}, and the projection function is chosen as a 2-layer MLP. After unsupervised training, a logistic regression classifier is applied to the learned embeddings for the downstream node classification task, where the final testing accuracy is selected at the epoch with the best validation accuracy. Experiments are repeated 20 times with accuracy reported by mean and standard deviation in Table~\ref{tab Node Classification}. Detailed model settings are summarized in Table~\ref{tab Node stats}. 

The results show that, except for Citeseer, iGCL outperforms classical node-level GCL methods DGI, GMI and GRACE on the rest five datasets. Compared to GCA that uses adaptive augmentation, iGCL performs better on Cora and Computers, and comparably on PubMed with a slight disadvantage.

\begin{figure*}[ht!]
\centering
\includegraphics[width=0.99\textwidth]{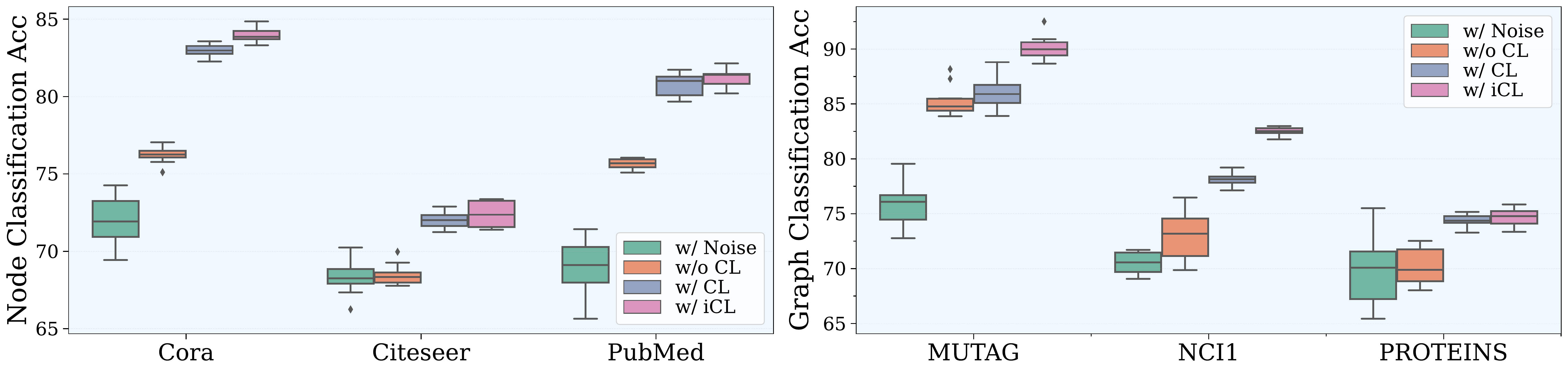} 

\caption{The ablation study of training with different modules at both node-level (left) and graph-level (right).}
\label{fig ablation}

\end{figure*}

\begin{figure*}[ht!]
\centering
\includegraphics[width=0.99\textwidth]{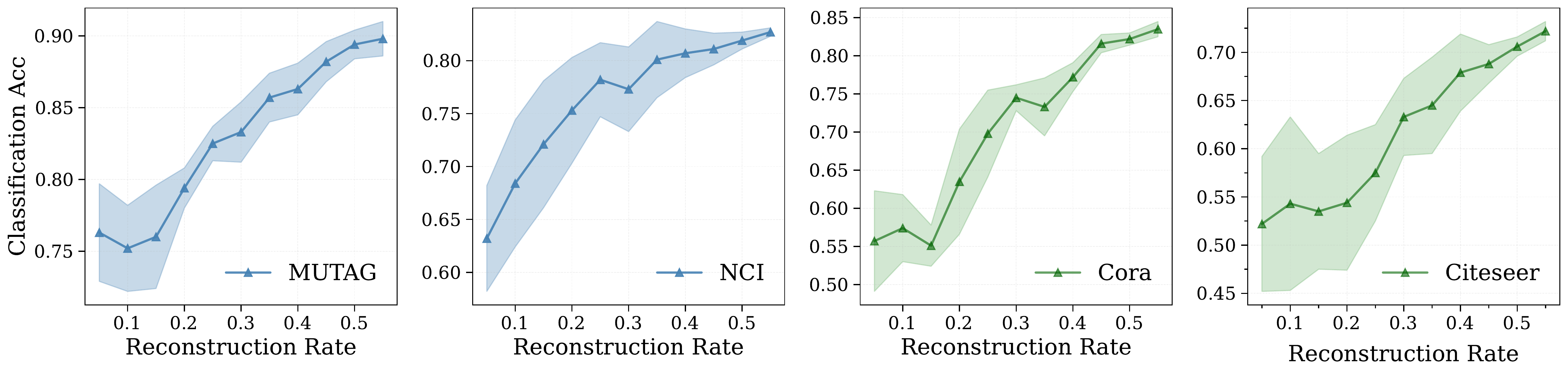} 
\caption{The sensitivity of graph (left two plots in blue) and node (right two plots in green) classification accuracy to reconstruction accuracy. Triangle represents the mean accuracy and shaded area represents the interval within one standard deviation. }

\label{fig adj acc}

\end{figure*}

\subsubsection{Node Clustering}
After contrastive training, we perform clustering on the learned embeddings using K-means algorithm, where three metrics are used for evaluation: accuracy (Acc), normalized mutual information (NMI) and adjusted random index (ARI), with results presented in Table~\ref{tab Node Clustering}. We can see that iGCL achieves comparable results with other state-of-the-art models on Cora and Citeseer datasets, meanwhile slightly outperforms the other baselines on PubMed under all metrics.

\subsection{Ablation Studies}
We investigate three cases to demonstrate the advantages of adopting implicit augmentations for graph contrastive learning: 1) Comparison between different training strategies; 2) the sensitivity of the downstream classification performance to VGAE reconstruction rate; and 3) the visual comparison of the embeddings learned from VGAE and our iGCL.

\subsubsection{Effectiveness of Different Modules}
We compare four different training strategies on both node-level and graph-level tasks: (a) contrasting with random Gaussian Noise, that is, fixing $\bm \mu = {\bf 0}$ and $\bf \Sigma = {\bf I}$ in Eq.~\eqref{eq entropy-like form} (denoted as w/ Noise), (b) using VGAE embeddings without contrastive learning (w/o CL), (c) using standard graph contrastive learning approach in Eq.\eqref{eq CL}, where we substitute ${\bm a}_n^{m}$ with embeddings generated by the same GNN backbone using augmented graph views based on node and edge perturbation (w/ CL), and (d) using full iGCL model (w/ iCL). The experiments are repeated 10 times with results summarized in Figure~\ref{fig ablation}. The box plots manifest that by adopting implicit augmentations, iGCL consistently outperforms classical contrastive methods on five datasets except for Citeseer, where iGCL underperforms by a small margin. Interestingly, contrasting with a standard normal distribution $\mathcal{N}({\bf 0},{\bf I})$ can achieve around 70\% classification accuracy on all six datasets, which is comparable to VGAE on Citeseer, NCI1 and PROTEINS. We attribute this phenomenon to the standard normal prior choice in VGAE, as $\mathcal{L}_{VGAE}$ in Eq.~\ref{eq VGAE} regularizes the latent distributions to the prior.

\subsubsection{Sensitivity to Reconstruction Accuracy}
In this section, we analyze the sensitivity of iGCL's performance regarding to the reconstruction accuracy of VGAE. We first train an independent VGAE and stop the training process when reconstruction rate is at around 10\%,20\%,...,60\%, then use the learned latent distributions ($\bm \mu$ and $\bm \sigma$) as the source for latent augmentations in iGCL, and finally evaluate the classification accuracy based on the learned embeddings at each reconstruction rate. We repeat the experiment 10 times and illustrate the results in Figure~\ref{fig adj acc}. We can see that at both graph-level and node-level, there is a positive correlation between the reconstruction rate and final classification accuracy, which corroborates the claim that latent augmentations with better quality can improve the contrastive learning performance on downstream tasks.

\begin{figure}[t!]
\centering
\includegraphics[width=0.99\columnwidth]{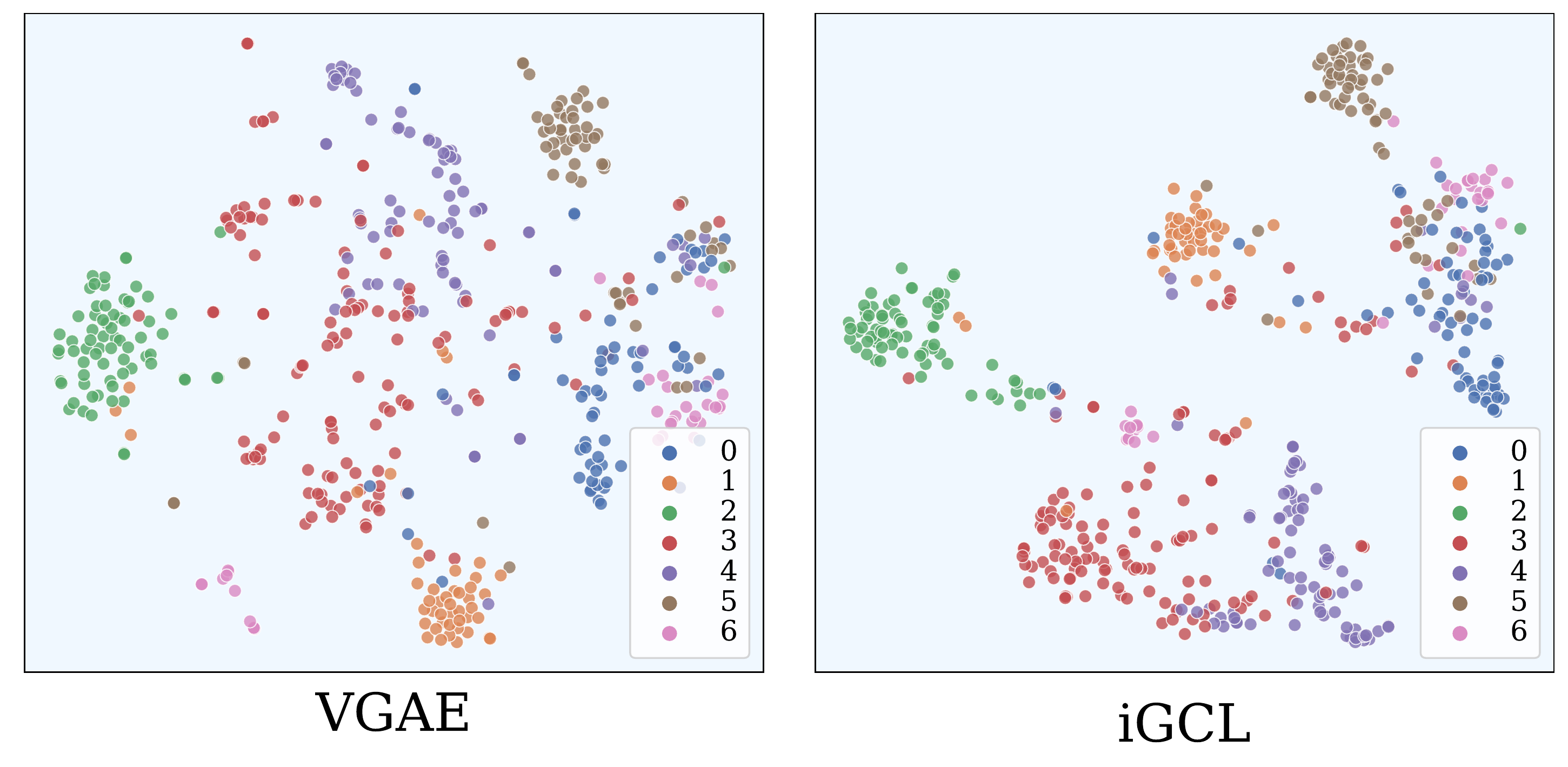} 
\caption{Embedding visualizations from VGAE (left) and iGCL (right) on Cora dataset using t-SNE, where colours represent node labels.}
\label{fig embedding}
\end{figure}

\subsubsection{Visual Comparison with VGAE embeddings}
To visualize the training results, we project the embeddings from VGAE and our iGCL into 2-dimensional space using t-SNE algorithm \cite{van2008visualizing} under the same settings on Cora dataset (50 perplexity with 1000 iterations), and present the embedding visualization in Figure~\ref{fig embedding}. While both methods can generate embeddings that are separable at some level, we can observe that nodes with red (3), purple (4) and blue (0) labels are more tightly clustered by iGCL as opposed to the results from VGAE, indicating the advantage of adopting implicit contrastive learning for downstream tasks.

\section{Conclusion}
In this paper we introduce \textit{Implicit Graph Contrastive Learning} (iGCL), an algorithm that exploits semantic augmentations in the latent space learned from a VGAE by reconstructing graph topological structure. The proposed method adopts an \textit{Implicit Contrastive Loss} that considers the expected contrastive loss w.r.t. all possible latent augmentations and is optimized by an entropy-like upper bound to improve computational efficiency. Experiments on both node-level and graph-level tasks indicate that iGCL achieves state-of-the-art performances compared to other GCL methods. Lastly, the ablation studies demonstrate the effectiveness of implicit contrastive learning and latent augmentation, which have potential extension to other machine learning domains.

\bibliography{ref}

\end{document}